%% file: emnlp2021.tex
\pdfoutput=1
\documentclass[11pt]{article}

\usepackage[]{emnlp2021}

\usepackage{times}
\usepackage{latexsym}

\usepackage[T1]{fontenc}
\usepackage[utf8]{inputenc}

\usepackage{microtype}      % make text margins prettier
\usepackage{graphicx}       % import graphical images
\usepackage[T1]{fontenc}    % fix font related glitches such as "|"
\usepackage{multirow}       % merge rows in tables
\usepackage{subcaption}     % create sub-tables and sub-figures
\usepackage{amssymb}        % enable \mathbb, \mathcal
\usepackage{bold-extra}     % enable \texttt{\textbf{}}
\usepackage{bm}             % enable bold font in math mode
\usepackage[hang,flushmargin]{footmisc}  % minimize footnote indentation
\newcommand{\LN}{\linebreak\noindent}    % to manage inline spacing
\usepackage{amsfonts,amsmath,amssymb}
\usepackage{array,booktabs}
\usepackage{enumitem}
\usepackage{stfloats}  % for bottom-aligned table
\usepackage{hyperref}

\title{Boosting Cross-Lingual Transfer via\\ Self-Learning with Uncertainty Estimation}

\author{
  Liyan Xu\textsuperscript{$\dagger$} \quad
  Xuchao Zhang\textsuperscript{$\ddagger$} \quad
  Xujiang Zhao\textsuperscript{$\mathsection$} \\
  \textbf{Haifeng Chen\textsuperscript{$\ddagger$} \quad
  Feng Chen\textsuperscript{$\mathsection$} \quad
  Jinho D. Choi\textbf{\textsuperscript{$\dagger$}}} \\
  \textsuperscript{$\dagger$}Emory University,  \{\texttt{liyan.xu,jinho.choi}\}\texttt{@emory.edu} \\
  \textsuperscript{$\ddagger$}NEC Laboratories America, \{\texttt{xuczhang,haifeng}\}\texttt{@nec-labs.com} \\
  \textsuperscript{$\mathsection$}University of Texas at Dallas, \{\texttt{xujiang.zhao,feng.chen}\}\texttt{@utdallas.edu}}

\begin{document}
\maketitle

\input{tex/abstract}
\input{tex/introduction}
\input{tex/related-work}
\input{tex/approach}
\input{tex/experiments}
\input{tex/conclusion}

\bibliography{references}
\bibliographystyle{acl_natbib}

\clearpage
\appendix
\input{tex/appendix}

\end{document}

%% file: tex/abstract.tex
\begin{abstract}
% Recent multilingual pre-trained language models have achieved remarkable zero-shot performance on cross-lingual transfer tasks, where the model is only finetuned on one source language and directly evaluated on multiple target languages. We present a simple framework that further utilizes unlabeled data of target languages, where the prediction of unlabeled data is used as silver labels in a self-training process. Uncertainty estimation is also combined within this process to select confident prediction. We evaluate our approach on cross-lingual Natural Language Inference and Named Entity Recognition, significantly improving the performance over the baseline. Our analysis shows the impact to the final performance of different uncertainty estimation, and the limitation of different uncertainty metrics.

Recent multilingual pre-trained language models have achieved remarkable zero-shot performance, where the model is only finetuned on one source language and directly evaluated on target languages.
In this work, we propose a\LN self-learning framework that further utilizes unlabeled data of target languages, combined with uncertainty estimation in the process to\LN select high-quality silver labels.
Three different uncertainties are adapted and analyzed specifically for the cross lingual transfer: Language Heteroscedastic/Homoscedastic Uncertainty (LEU/LOU), Evidential  Uncertainty (EVI).
We evaluate our framework with uncertainties on two cross-lingual tasks including Named Entity Recognition (NER) and Natural Language Inference (NLI) covering 40 languages in total, which outperforms the baselines significantly by 10 F1 on average for NER and 2.5 accuracy score for NLI.

\end{abstract}

%% file: tex/introduction.tex
\section{Introduction}
\label{sec:introduction}

Recent multilingual pre-trained language models such as mBERT \citep{bert}, XLM-R \citep{xlmr} and mT5 \citep{mt5} have demonstrated remarkable performance on various direct zero-shot cross-lingual transfer tasks, where the model is finetuned on the source language, and directly evaluated on multiple target languages that are unseen in the task-finetuning stage.
While direct zero-shot transfer is a sensible testbed to assess the multilinguality of language models, one would apply supervised or semi-supervised learning on target languages to obtain more robust and accurate predictions in a practical scenario.

In this work, we investigate self-learning (also known as ``pseudo labels'') as one way to apply semi-supervised learning on cross-lingual transfer, where only unlabeled data of target languages are required, without any efforts to annotate gold labels for target languages.
As self-learning has been proven effective in certain tasks of computer vision \citep{sl1,sl4} and natural language processing \citep{sl5,sl2,sl6}, we propose to formalize an iterative self-learning framework for multilingual tasks using pre-trained models, combined with explicit uncertainty estimation in the process to guide the cross-lingual transfer.

Our self-learning (SL) framework utilizes any multilingual pre-trained models as the backbone, and iteratively grows the training set by adding predictions of target language data as silver labels.
We reckon two important observations from our preliminary study (baselines in \textsection\ref{sec:experiments}).
First, compared with self-training one target language at a time, jointly training multiple languages together can improve the performance on most languages, especially for certain low-resource languages that can achieve up to 8.6 F1 gain in NER evaluation.
Therefore, our SL framework features the joint training strategy, maximizing potentials of different languages benefiting each other.
Second, compared with simply using all unlabeled data as silver labels without considering prediction confidence, estimating uncertainties becomes critical in the transfer process,\LN as higher quality of silver labels should lead to better performance.
We hence introduce three different uncertainty estimations in the SL framework.

Specifically, we adapt uncertainty estimation techniques based on variational inference and evidence learning for our cross-lingual transfer, namely LEU, LOU and EVI (\textsection\ref{subsec:un}).
We evaluate our framework and three uncertainties on two multilingual tasks from XTREME \citep{xtreme}: Named Entity Recognition (NER), and Natural Language Inference (NLI).
Empirical results suggest LEU to be the best uncertainty estimation overall, while the others can also perform well on certain languages (\textsection\ref{subsec:results}).
Our analysis shows further evaluation of different estimations, corroborating the correlation between the uncertainty quality and the final SL performance.
Characteristics of different estimations are also discussed, including the language similarities learned by LOU and the current limitation of EVI in the SL process (\textsection\ref{sec:analysis}).

Our contributions in this work can be summarized as follows.
(1) We propose the self-learning framework for the cross-lingual transfer and identify the importance of uncertainty estimation under this setting.
(2) We adapt three different uncertainty estimations in our framework, and evaluate the framework on both NER and NLI tasks covering 40 languages in total, improving the performance of both high-resource and low-resource languages on both tasks by a solid margin (10 F1 for NER and 2.5 accuracy score for NLI on average).
(3) Further analysis is conducted to compare different uncertainties and their characteristics.

%% file: tex/related-work.tex
\section{Related Work}
\label{sec:related-work}

We introduce the work of uncertain estimation briefly. As deep learning models are optimized by minimizing the loss without special care on the uncertainty, they are usually poor at quantifying uncertainty and tend to make over-confident predictions, despite producing high accuracies \citep{over_conf}.
Estimating the uncertainty of deep learning models has been recently studied in NLP tasks \citep{xiao2019quantifying, zhang2019mitigating, he2020towards}.
There are two main uncertainty types in Bayesian modelling \citep{kendall17, pmlr-v80-depeweg18a}: \textit{epistemic} uncertainty that captures the model uncertainty itself, which can be explained away with more data; \textit{aleatoric} uncertainty that captures the intrinsic data uncertainty regardless of models.
\textit{Aleatoric} uncertainty can further be devided into two sub-types: \textit{heteroscedastic} uncertainty that depends on input data, and \textit{homoscedastic} uncertainty that remains constant for all data within a task.
In this work, we only focus on \textit{aleatoric} uncertainty, as it is more closely related to our SL process to select confident and high-quality predictions within each iteration.

%% file: tex/approach.tex
\section{Approach}
\label{sec:approach}

We keep the same model architecture throughout our experiments:
a multilingual pre-trained language model is employed to encode each input sequence, followed by a linear layer to classify on the hidden state of CLS token for NLI, and of each token for NER, which is the same model setting from XTREME \citep{xtreme}.
Cross-entropy (CE) loss is used during training in the baseline.

\begin{figure}[t]
\centering
\includegraphics[width=0.75\columnwidth]{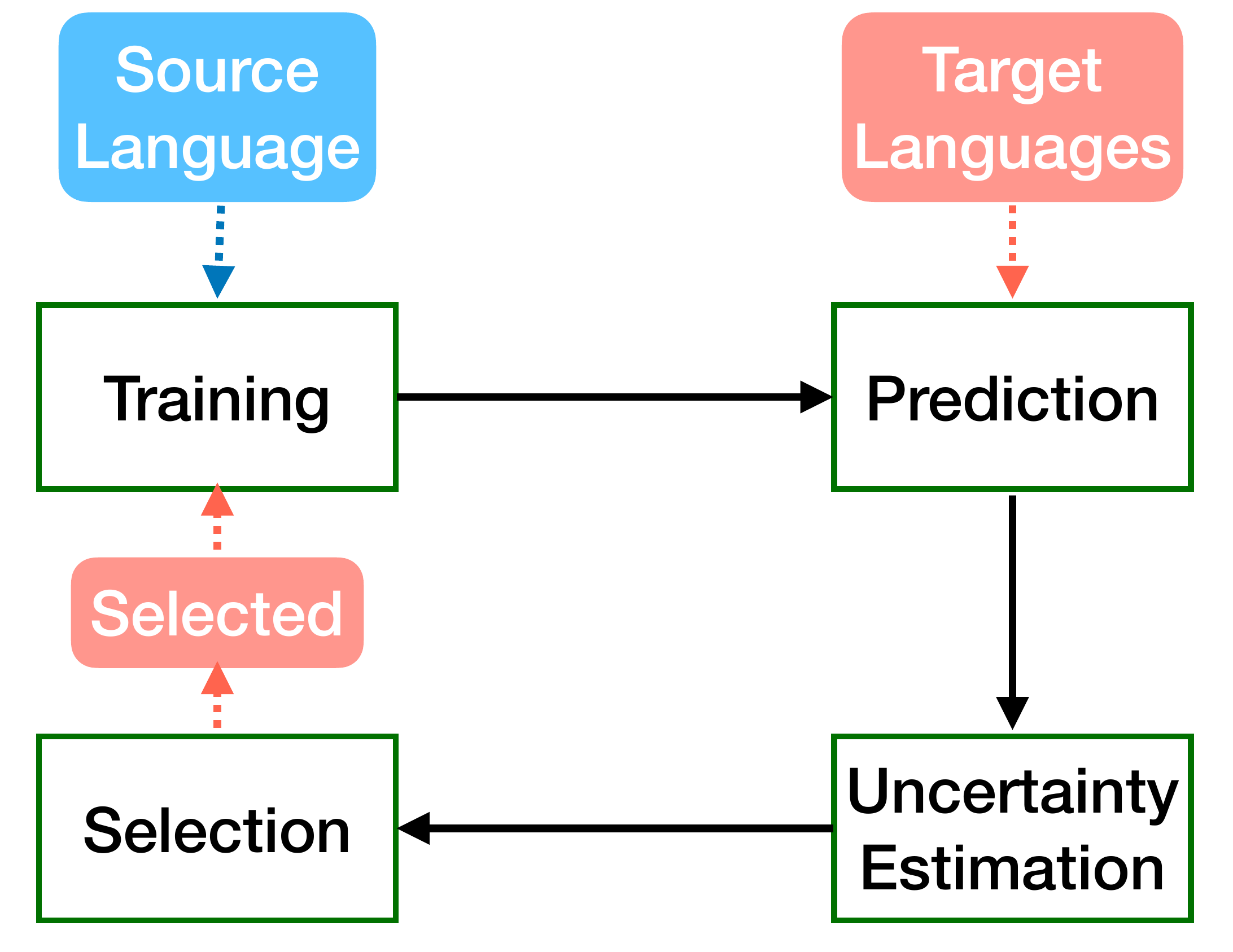}
\caption{Illustration of the self-learning framework with explicit uncertainty estimation.}
\label{fig:sl}
\vspace{-2ex}
\end{figure}

\subsection{Self-Learning (SL) Framework}
\label{subsec:sl}

We formulate the task-agnostic SL framework for cross lingual transfer into the following four phases, as shown in Figure~\ref{fig:sl}.
In the \textit{training} phase, the model parameter $\theta$ gets optimized by the training inputs $X$ and labels $Y$, with $Y$ being gold labels of the source language in the first iteration, along with silver labels of target languages in later iterations.
Inputs of different languages are mixed together.
In the \textit{prediction} phase, the model predicts on the remaining unlabeled data $X^*_l = \{x^*_{l1},\dots,x^*_{lN}\}$ of each target language $l$, with each prediction denoted as $y^* = f^{\theta} (x^*)$.
In the \textit{uncertainty estimation} phase, the model estimates the prediction uncertainty based on one of the methods described in \textsection\ref{subsec:un}, denoted as $\gamma = f^{\theta}_{\gamma} (x^*, y^*)$, representing the model confidence of the prediction.
In the \textit{selection} phase, data in each $X^*_l$ is ranked based on the uncertainty score $\gamma$, and we select top-K percent of each $X^*_l$ with their predictions as silver labels, adding to the training data.
To avoid posing potential inductive bias from imbalanced label distribution, we select equal amount of inputs for each label type, similar to previous\LN work on self-learning \citep{sl1,sl2,sl3}.

After \textit{selection}, the model goes back to the \textit{training} phase and starts a new iteration with the updated training set.
The entire process keeps iterating until there is no remaining unlabeled data; early stop criteria are implemented on the dev set of the source language only, as gold labels are not available for other languages. Each phase can be adjusted by task-specific requirements (see \ref{apdx:adjustment}).

\subsection{Uncertainty Estimation}
\label{subsec:un}

We adapt three different uncertainty estimation techniques in our framework. Let $C$ be the label classes, $p_c$ be the probability of class $c$ for an input.

\paragraph{Language Heteroscedastic Uncertainty (LEU)} LEU injects Gaussian noise into class logits whose variance is predicted by the model as an input-dependent uncertainty \citep{kendall17}, regardless of languages.
A Gaussian distribution is placed on the logit space $\mathbf{g} \sim \mathcal{N}(g^{\theta}, (\sigma^{\theta})^2)$, where the model is modified to predict both raw logit $g^{\theta}$ and standard deviation $\sigma^{\theta}$ given each input.
We use the expectation of the logit softmax as the new probability, computed by Monte Carlo sampling: $p_c = E [\text{softmax} (\mathbf{g}_c)] \approx \frac{1}{T} \sum_t \text{softmax} (g_{tc})$, with $g_{tc}$ being the logit of class $c$ at $t$-th sampling from $\mathbf{g}$.
The training loss and the uncertainty take into account the new probability formulation $p_c$:
\begin{align}
\label{eq:var}
    \mathcal{L}^{\text{LEU}} = -\log \frac{1}{T} \sum_t \exp \big( -\mathcal{L}_t (x, c) \big)
\end{align}
The loss is composed of the CE loss $\mathcal{L}_t (x, c)$ on input $x$ and gold class $c$ with $t$th sampled probabilities.
The uncertainty is the entropy of the new probabilities: $\gamma = -\sum_c p_c \log p_c$. When an input of any language is hard to predict, the model will signal high variance, indicating high uncertainty,\LN as the probability distribution tends to be uniform.

\paragraph{Language Homoscedastic Uncertainty (LOU)}
LOU estimates the uncertainty of each certain language, regardless of the input.
Similar to the formulation of task uncertainty \citep{multitask}, we propose to place an uncertainty $\sigma_l$ on a language $l$ as the \textit{homoscedastic} uncertainty.
$\sigma$ is used as the softmax temperature on the predicted logits $g^{\theta}$: $p_c = \text{softmax} (\frac{1}{\sigma^2}g^{\theta}_c)$.
The final uncertainty is also the entropy of the scaled probabilities.
A higher $\sigma_l$ leads to higher entropy of all inputs of language $l$, as the probability distribution tends to be more uniform.
During training, each $\sigma_l$ is a learned parameter directly, and the new loss for an input of language $l$ can be approximated as:
\begin{align}
\label{eq:lun}
    \mathcal{L}^{\text{LOU}} \approx \frac{1}{\sigma^2_l} \mathcal{L}(x,c) + \log \sigma_l
\end{align}
$\mathcal{L}(x,c)$ is the same CE loss as in Eq~\eqref{eq:var}.
Note that LOU does not change the input selection nor ranking within each language; we mainly use it as an optimization strategy to jointly train inputs of multiple languages, automatically distinguishing the importance of different target languages.

\paragraph{Evidential Uncertainty (EVI)}
EVI estimates the evidence-based uncertainty \citep{evi_un}, where the softmax probability is replaced with Dirichlet distribution, and each predicted logit for class $c$ is regarded as the evidence.
We employ the decomposed entropy \textit{vacuity} and \textit{dissonance} proposed by \citet{xujiang}.
\textit{vacuity} is high when there lacks evidence for all the classes, indicating out-of-distribution (OOD) samples that are far away from the source language; \textit{dissonance} becomes high when there are conflicts of strong evidence among certain classes (more details are shown in \ref{apdx:approach_un}).
The prediction is said uncertain if either \textit{vacuity} or  \textit{dissonance} is high.
For each input, let $S$ be the total evidence strength, and let the label $y_c$ be $1$ for the gold class and $0$ for the others. The following describes the expected probability $p_c$ for the class $c$ under Dirichlet distribution, as well as the training loss $\mathcal{L}^{\text{EVI}}$:
\begin{align}
    p_c &= \frac{e_c + 1}{S} \label{eq:evi_pc} \\
    \mathcal{L}^{\text{EVI}} &= \sum_c (y_c - p_c)^2 + \frac{p_c(1-p_c)}{S + 1} \label{eq:evi} 
\end{align}

%% file: tex/experiments.tex
\section{Experiments}
\label{sec:experiments}

The framework with different uncertainties are evaluated on two cross-lingual transfer datasets: XNLI \citep{xnli} for the NLI task covering 15 languages, and Wikiann \citep{panx} for the NER task covering 40 languages.
For both datasets, English is the source language with gold labels, and we use the dev set of target languages (TLs) as the source of unlabeled data; we do not consult any gold labels of TLs in the SL process.
XLM-R\textsubscript{Large} \citep{xlmr} is used as the multilingual encoder across our experiments.
Our detailed experimental setting can be found in \ref{apdx:experiments}.

We implement three different settings for the baseline.\footnote{Code is available at \href{https://github.com/lxucs/multilingual-sl}{https://github.com/lxucs/multilingual-sl}.}
\textbf{BL-Direct} is the direct zero-shot transfer without utilizing unlabeled data of TLs.
\textbf{BL-Single} trains gold data of English and silver data of only one TL per model; it simply selects predictions of all unlabeled data as silver labels, without considering any uncertainties.
\textbf{BL-Joint} is similar to BL-Single but instead train with all TLs jointly.

For SL, we set top-K percent selection to be top 8\% of total unlabeled data for each label type, so the entire SL process will finish in around 6 iterations.
We found that K below 10\% can generally yield decent performance.

For the analysis, we also include two common uncertainties used in previous work of self-learning on other tasks: max probability (MPR), and entropy (ENT); both use plain softmax probabilities (\ref{apdx:common}).
\input{tab/ner}
\input{tab/nli}

\subsection{Results}
\label{subsec:results}

The results for NER and NLI are shown in Table~\ref{tab:results_ner} and \ref{tab:results_nli} respectively.
BL-Direct is equivalent to our re-implementation of \citet{xtreme}.

BL-Single outperforms BL-Direct on NER by 3.1 F1 on average, demonstrating the effectiveness of utilizing unlabeled data even without considering uncertainties.
Remarkablely, languages such as Arabic (ar), Japanese (ja), Urdu (ur) and Chinese (zh) receive 10+ gain in F1.
By contrast, BL-Single does not surpass the baseline for NLI, partially because all TLs already have much closer performance to English, which in turn highlights the importance of estimating uncertainties for SL.

BL-Joint outperforms BL-Single on both tasks by a slight margin, and we do see performance gain over BL-Single on 32/40 and 10/15 languages for NER and NLI respectively.
Certain languages such as Hindi (hi), Javanese (jv) and Yoruba (yo) receive non-trivial benefits (2.6 - 8.6 F1 gain for NER) through the joint language training, validating our joint training strategy for SL.

Evaluation of SL is shown with the best results of each uncertainty from 3 repeated runs.
The best performance of SL for both tasks is achieved by adopting LEU as the uncertainty estimation, which outperforms three baselines significantly (10\% gain for NER and 2.5\% for NLI on average), and surpasses other uncertainties by a slight margin.
In NER specifically, certain low-resource languages such as Basque (eu), Persian (fa), Burmese (my) and Urdu (ur) have substantial performance improvement over BL-Joint (13.8 - 25.9 F1 gain);
the performance of certain high-resource languages such as Arabic (ar), German (de) and Chinese (zh) can also increase by a solid margin over BL-Joint (4.1 - 13.3 F1 gain). The trend of improving both high and low-resource languages is also present in NLI.
All results are stable across multiple runs with standard deviation within 0.1 - 0.2 on average.

Results also suggest that other uncertainty estimations can achieve comparable performance, as LEU does not dominate every language. We further conduct analysis on uncertainties as follows.

\section{Analysis}
\label{sec:analysis}

\paragraph{Uncertainty Comparison}
To directly assess different uncertainty estimations, we evaluate uncertainty scores by AUROC against predictions, such that AUROC is high when the model is confident on correct predictions and uncertain on incorrect predictions.
The left side of Table~\ref{tab:analysis} shows the AUROC of four estimations on the test sets of both tasks. MPR and ENT are also included in the experiments for comparison; LOU is excluded as it does not change selection.
The right side of Table~\ref{tab:analysis} shows the SL performance drop using other uncertainties compared to LEU, serving as an indirect evaluation of different uncertainties.
As shown, LEU indeed achieves the best AUROC, being a better uncertainty estimation compared to others;
EVI has the lowest AUROC and also the lowest SL performance;
MPR and ENT can bring moderate scores on both AUROC and SL.
Thus, Table~\ref{tab:analysis} corroborates strong correlation between AUROC and SL performance: better uncertainty can indeed lead to higher performance in the SL process.

\begin{table}[t!]
%\small
\centering
\resizebox{\columnwidth}{!}{
\begin{tabular}{l|cccc|cccc}
% \toprule
& \tt M & \tt T & \tt I & \tt E & \tt M & \tt T & \tt I & \tt O \\
\midrule
NER & 71.2 & 72.1 & 68.7 & \bf 73.7 & 0.6 & 0.5 & \bf 1.1 & 0.5 \\
XNLI & 76.9 & 77.3 & 73.0 & \bf 78.6 & 0.3 & 0.3 & \bf 0.7 & 0.0 \\
% \bottomrule
\end{tabular}}
\vspace{-1ex}
\caption{The left side shows the averaged AUROC of different uncertainty estimations. The right side shows the averaged SL performance drop compared to LEU.  \texttt{M}=MPR, \texttt{T}=ENT, \texttt{I}=EVI, \texttt{O}=LOU, \texttt{E}=LEU.}
\label{tab:analysis}
\end{table}

\paragraph{Language Uncertainty}
Table~\ref{tab:results_nli} shows that LOU reaches the same accuracy as LEU on XNLI, with trivial performance gap for each language.
We find that the learned uncertainty of each language is highly consistent through multiple runs, as shown in Table~\ref{tab:results_nli_lou}, which can be loosely interpreted as language similarities under the input of this task, e.g. Vietnamese (vi) appears to be more distant from English than others for this task, and the joint optimization of all languages could benefit from this learned language uncertainty.
However, we do not find LOU to be as stable on NER, potentially because NER has much more noise and languages.

\input{tab/nli_lou}

\begin{figure}[ht]
\centering
\includegraphics[width=0.7\columnwidth]{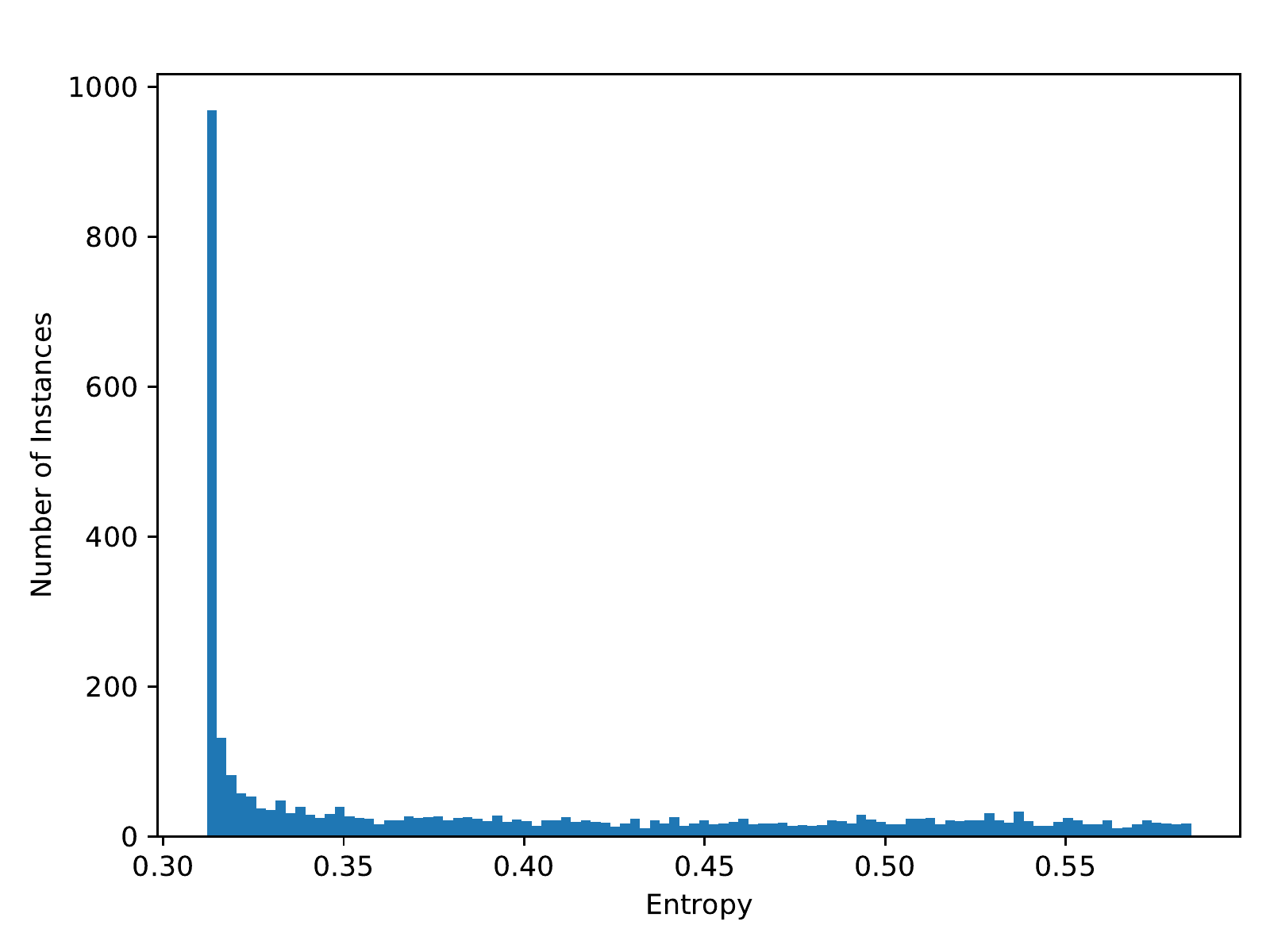}
\caption{Evidence-based entropy distribution on the test set of NER for Japanese (ja).}
\label{fig:dist}
\vspace{-2ex}
\end{figure}

\begin{figure}[ht]
\centering
\includegraphics[width=\columnwidth]{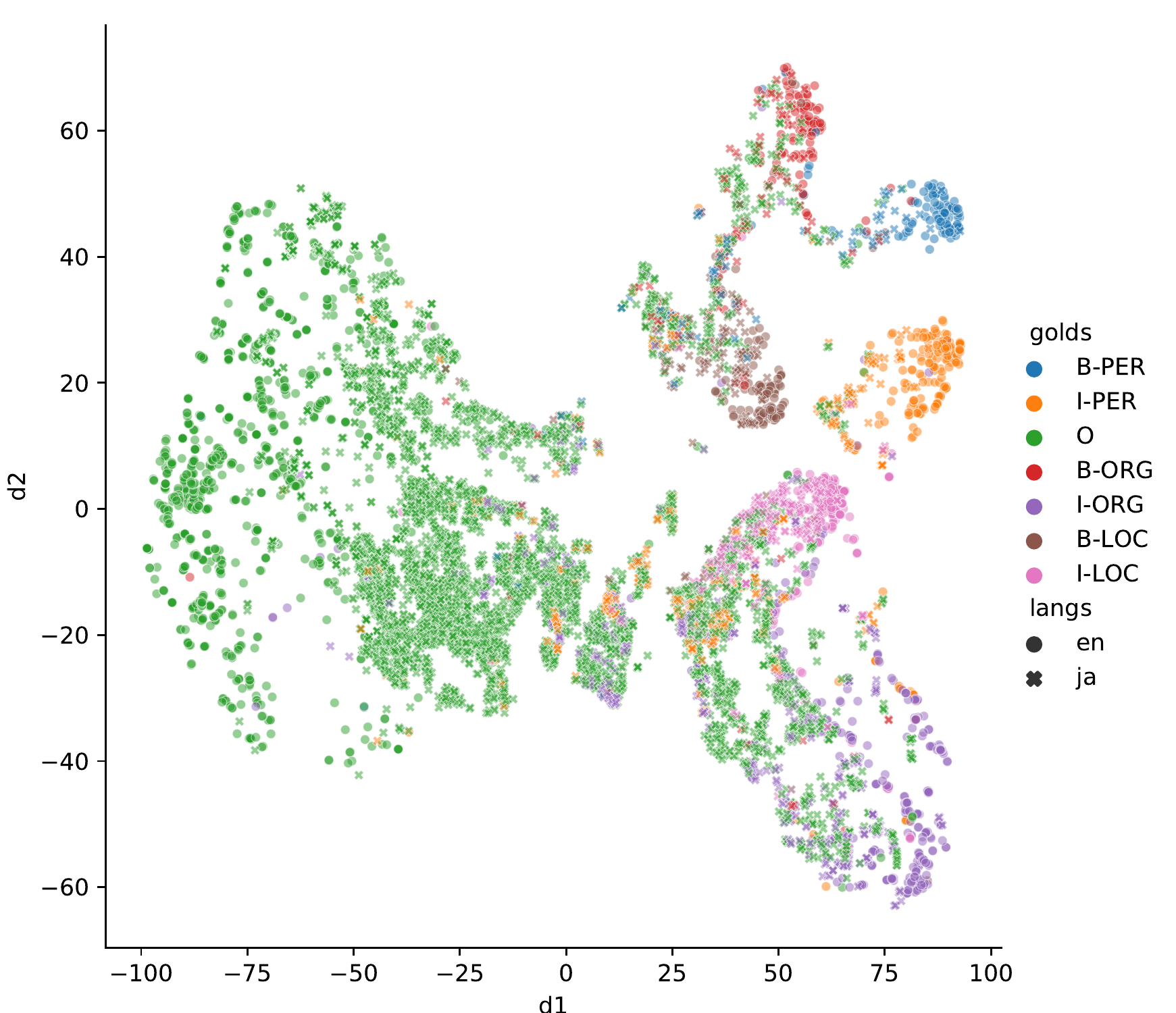}
\caption{t-SNE visualization of CLS hidden states of NER inputs in English (en) and Japanese (ja). Different gold label types for tokens are marked by colors, and two languages are marked by the shapes. Each cluster should ideally has only one distinct color.}
\label{fig:ood}
\vspace{-2ex}
\end{figure}

\paragraph{Evidential Uncertainty}
Although EVI is able to achieve good performance on certain languages, there also exists large gap for certain other languages compared to LEU.
We attribute the inferior performance of EVI to two aspects.
First, the predicted evidence (logit) still exhibits over-confidence, which destabilizes the \textit{vacuity} and \textit{dissonance}. Figure~\ref{fig:dist} shows an example of the evidence-based entropy distribution for EVI, and the model indicates most all predictions as certain (small entropy).
% Regularization in training is needed as future work.
Second, \textit{vacuity} can only distinguish true OOD samples for English, as only English has gold labels.
It could fail to recognize those confident samples of TLs that appear in-distribution but are inherently wrong, and falsely select them in the SL process. Figure~\ref{fig:ood} shows the t-SNE visualization of hidden states of inputs in English and Japanese on the test set of NER: some target language inputs that are close to English in terms of hidden states are predicted wrong, because of the zero-shot nature.

%% file: tab/ner.tex
\begin{table*}[ht!]
%\small
\centering
\resizebox{\textwidth}{!}{
\begin{tabular}{l|cccccccccccccccccccc|c}
% \toprule
\midrule
& en & af & ar & bg & bn & de & el & es & et & eu & fa & fi & fr & he & hi & hu & id & it & ja & jv \\
\midrule
% \citet{xtreme} & 84.7 & 78.9 & 53.0 & 81.4 & 78.8 & 78.8 & 79.5 & 79.6 & 79.1 & 60.9 & 61.9 & 79.2 & 80.5 & 56.8 & 73.0 & 79.8 & 53.0 & 81.3 & 23.2 & 62.5 \\
BL-Direct & 84.0 & 79.3 & 45.5 & 81.4 & 77.4 & 78.8 & 78.9 & 71.4 & 79.0 & 61.0 & 52.0 & 78.7 & 79.3 & 54.6 & 70.8 & 79.4 & 52.9 & 81.0 & 25.0 & 62.6 \\
BL-Single & 84.0 & 78.9 & 56.9 & 84.5 & 79.3 & 80.9 & 81.6 & 72.9 & 80.7 & 63.2 & 54.8 & 80.5 & 81.9 & 63.0 & 73.9 & 81.7 & 54.3 & 82.1 & 36.5 & 60.9 \\
BL-Joint & 84.7 & 79.5 & 56.7 & 84.9 & 80.5 & 80.5 & 81.5 & 73.3 & 81.2 & 64.0 & 55.1 & 81.2 & 82.1 & 62.6 & 76.6 & 81.6 & 54.5 & 83.0 & 37.2 & 63.5 \\
\midrule
SL-EVI & \bf 85.2 & 83.7 & \bf 75.1 & 85.8 & 82.0 & 83.6 & 84.4 & \bf 86.5 & 84.6 & 72.1 & 72.9 & 84.7 & \bf 84.1 & 61.4 & \bf 80.2 & 85.7 & 54.8 & 83.9 & 41.3 & 69.2 \\
SL-LOU & 84.4 & \bf 85.3 & 61.1 & 87.1 & 81.9 & 83.4 & 85.4 & 75.6 & 85.5 & 74.6 & 74.9 & 84.4 & 83.3 & \bf 68.5 & 78.6 & 84.5 & \bf 55.5 & \bf 85.1 & 46.2 & \bf 70.0 \\
SL-LEU & 84.7 & 81.5 & 70.0 & \bf 87.6 & \bf 83.6 & \bf 84.6 & \bf 85.5 & 85.0 & \bf 85.6 & \bf 77.8 & \bf 81.0 & \bf 86.2 & 83.1 & 62.0 & 79.5 & \bf 87.0 & 53.4 & 84.8 & \bf 49.5 & 65.3 \\
\bottomrule
& ka & kk & ko & ml & mr & ms & my & nl & pt & ru & sw & ta & te & th & tl & tr & ur & vi & yo & zh & avg \\
\midrule
% \citet{xtreme} & 71.6 & 56.2 & 60.0 & 67.8 & 68.1 & 57.1 & 54.3 & 84.0 & 81.9 & 69.1 & 70.5 & 59.5 & 55.8 & 1.3 & 73.2 & 76.1 & 56.4 & 79.4 & 33.6 & 33.1 & 65.4 \\
BL-Direct & 69.3 & 51.9 & 57.9 & 63.6 & 62.4 & 69.6 & 60.1 & 83.7 & 80.9 & 70.2 & 69.2 & 58.2 & 51.3 & 1.8 & 71.0 & 76.7 & 55.8 & 76.2 & 41.4 & 33.0 & 64.4 \\
BL-Single & 73.6 & 52.5 & 63.6 & 66.0 & 66.8 & 62.6 & 54.3 & 84.8 & 82.6 & 72.9 & 67.7 & 63.2 & 57.2 & 3.1 & 74.7 & 81.8 & 69.9 & 80.9 & 46.2 & 43.6 & 67.5 \\
BL-Joint & 73.6 & 53.4 & 63.6 & 67.5 & 67.9 & 64.3 & 53.0 & 84.8 & 83.2 & 73.5 & 69.7 & 63.1 & 57.4 & 3.6 & 76.1 & 81.8 & 71.5 & 81.4 & \bf 54.8 & 43.7 & 68.3 \\
\midrule
SL-EVI & 81.0 & 56.4 & 69.4 & \bf 76.3 & 77.9 & 72.5 & 71.7 & \bf 87.1 & 85.5 & \bf 80.6 & 71.2 & 69.4 & 61.5 & \bf 6.7 & 80.7 & 85.3 & 79.8 & \bf 86.2 & 42.7 & 48.9 & 73.3 \\
SL-LOU & 78.8 & 58.7 & 70.2 & 75.4 & \bf 79.4 & 73.8 & 71.2 & 86.4 & 86.2 & 79.2 & \bf 73.3 & \bf 69.5 & 68.8 & 4.7 & \bf 83.4 & 88.4 & \bf 85.9 & 85.8 & 49.1 & 50.5 & 73.8\\
SL-LEU & \bf 81.1 & \bf 63.7 & \bf 71.8 & 76.0 & 76.2 & \bf 75.9 & \bf 71.5 & \bf 87.1 & \bf 87.6 & 79.9 & 70.4 & 64.0 & \bf 69.9 & 2.2 & 81.3 & \bf 89.1 & \bf 85.9 & 85.9 & 43.5 & \bf 54.8 & \bf 74.4 \\
\bottomrule
\end{tabular}}
\vspace{-1ex}
\caption{NER Results in F1 scores for 40 languages. BL-Direct is equivalent to \citet{xtreme}.}
\label{tab:results_ner}
\end{table*}

%% file: tab/nli.tex
\begin{table*}[ht!]
%\small
\centering
\resizebox{0.9\textwidth}{!}{
\begin{tabular}{l|ccccccccccccccc|c}
% \toprule
\midrule
& en & ar & bg & de & el & es & fr & hi & ru & sw & th & tr & ur & vi & zh & avg \\
\midrule
% \citet{xtreme} & \bf 88.7 & 77.2 & 83.0 & 82.5 & 80.8 & 83.7 & 82.2 & 75.6 & 79.1 & 71.2 & 77.4 & 78.0 & 71.7 & 79.3 & 78.2 & 79.2 \\
BL-Direct & 88.5 & 78.0 & 82.5 & 81.8 & 80.5 & 83.8 & 82.9 & 74.8 & 78.7 & 67.5 & 76.7 & 78.1 & 71.5 & 79.4 & 78.2 & 78.9 \\
BL-Single & 88.5 & 77.6 & 82.4 & 82.0 & 79.6 & 82.5 & 82.1 & 76.1 & 79.1 & 69.1 & 76.6 & 77.9 & 71.5 & 77.9 & 78.2 & 78.7 \\
BL-Joint & 88.2 & 78.8 & 82.0 & 82.2 & 80.4 & 83.1 & 82.2 & 76.1 & 79.6 & 68.8 & 76.2 & 78.0 & 71.4 & 79.1 & 78.5 & 79.0 \\
\midrule
SL-EVI & 88.1 & 79.5 & 84.4 & 83.4 & 82.4 & \bf 84.8 & 83.7 & 78.0 & 81.6 & 71.1 & 78.2 & 79.2 & 74.4 & 80.8 & 80.4 & 80.7 \\
SL-LOU & 88.2 & \bf 81.0 & 84.4 & \bf 83.5 & 82.3 & \bf 84.8 & \bf 83.9 & 78.9 & \bf 81.8 & \bf 73.9 & 79.3 & 80.1 & \bf 75.7 & 81.6 & \bf 81.4 & \bf 81.4 \\
SL-LEU & 88.1 & 80.7 & \bf 84.9 & 83.4 & \bf 82.8 & 84.5 & 83.8 & \bf 79.2 & \bf 81.8 & 73.0 & \bf 79.7 & \bf 80.5 & \bf 75.7 & \bf 81.9 & 81.3 & \bf 81.4 \\
\bottomrule
\end{tabular}}
\vspace{-1ex}
\caption{XNLI Results in accuracy scores for 15 languages.}
\label{tab:results_nli}
\end{table*}

%% file: tab/nli_lou.tex
\begin{table}[ht]
%\small
\centering
\resizebox{0.92\columnwidth}{!}{
\begin{tabular}{cccccccc}
\toprule
 en & ar & bg & de & el & es & fr & hi \\
\midrule
 1.44 & 1.20 & 1.15 & 0.63 & 0.58 & 1.78 & 0.70 & 1.60 \\
\midrule
\midrule
ru & sw & th & tr & ur & vi & zh \\
\midrule
0.33 & 1.07 & 4.18 & 1.89 & 3.15 & 0.23 & 0.99 \\
\bottomrule
\end{tabular}}
\vspace{-1ex}
\caption{The learned language uncertainty $\sigma^2$ of LOU for each language in XNLI.}
\label{tab:results_nli_lou}
\end{table}

%% file: tex/conclusion.tex
\section{Conclusion}
\label{sec:conclusion}

In this work, we propose a self-learning framework combined with explicit uncertainty estimation for cross-lingual transfer.
% which only requires unlabeled data of target languages.
Three different uncertainties are adapted, and the entire framework is evaluated on two tasks of NER and NLI, surpassing the baseline by a large margin.
Further analysis shows the evaluation and characteristics of each uncertainty.

%% file: tex/appendix.tex
\section{Appendix}
\label{sec:appendix}

% \subsection{Self-Learning Framework}
% \label{apdx:sl}

% Figure~\ref{fig:sl} shows the overall process of our self-learning framework that consists of four phases.
% \begin{figure}[h]
% \centering
% \includegraphics[width=0.8\columnwidth]{fig/sl.png}
% \caption{The self-learning framework w/ uncertainty.}
% \label{fig:sl}
% \vspace{-2ex}
% \end{figure}
% % TODO: this is a placeholder

\subsection{Uncertainty Estimation}
\label{apdx:approach_un}

For \textbf{LOU}, the uncertainty term as the denominator in the loss as in Eq~\eqref{eq:lun} achieves the effect of ``learned loss attenuation'' \citep{kendall17} during training, where uncertain samples have lower scale of loss, so that the optimization is less prone to noisy data. We use LOU to let the model learn the uncertainty for each language to achieve more stable training amid selected data with silver labels.

In practice, the model directly predicts the log-variance term $\log \sigma$ for both LEU and LOU, as the training is more stable and the variance is guaranteed to be positive.

\noindent For \textbf{EVI}, we follow \citet{evi_un} and define:
\begin{align}
    b_c = \frac{e_c}{S} \quad u = \frac{|C|}{S} \quad S = \sum_c e_c + |C|
\end{align}
$e_c$ is the evidence strength (logit) for class $c$, $|C|$ is the number of classes. $b_c$ represents the belief mass for class $c$ and $u$ is the \textit{vacuity}, denoted as $vac = u$. We follow \citet{xujiang} and define \textit{dissonance} for each input as:
\begin{align}
    \text{Bal} (b_j,b_k) &=
    \begin{cases}
    1 - \frac{|b_j - b_k|}{b_j + b_k} \quad \text{if } b_i b_j \neq 0\\
    0  \hfill \text{elsewise}
    \end{cases}\\
    diss &= \sum_c \frac{b_c \sum_{c' \neq c} b_{c'} \text{Bal} (b_c, b_{c'})}{\sum_{c' \neq c} b_{c'}}
\end{align}
Both \textit{vac} and \textit{diss} are in the range of $[0, 1]$; being closer to $1$ indicates more uncertainty.
The final uncertainty is set as $\gamma = diss + \alpha \cdot vac$ with $\alpha$ being a hyperparameter.

In practice, ELU activation is added after raw logits to ensure the evidence strength is positive.

\subsection{Task-Specific Adjustment}
\label{apdx:adjustment}

We adjust the SL process for NER as follows: the uncertainty score is obtained for each predicted entity, which is calculated as the averaged uncertainty score of all tokens within the entity.
Ranking is performed on entities within each entity type; we select the input sequence if all its predicted entities have uncertainty within the top-K threshold.

\subsection{Experimental Setting}
\label{apdx:experiments}

We follow the same train/dev/test split and same evaluation protocol as XTREME \citep{xtreme}.

\paragraph{Datasets}
For XNLI \citep{xnli}, there are three label types for each sequence: ``neutral'', ``entailment'', ``contradiction''. For Wikiann \citep{panx}, there are three entity types: ``LOC'', ``PER'', ``ORG''; each token is tagged in the BIO2 format, thus there are 7 label types for each token.

\paragraph{Hyperparameters}
For both NLI and NER, we use the following hyperparameter setting as suggested by XTREME \citep{xtreme}: 32 effective batch size, $2 \times 10^{-5}$ learning rate with linear decay scheduling,  $1$ max gradient norm.

For NLI in the self-learning (SL) process, we train the model by 5 epochs in the first iteration on English training set with gold labels, whereas we train 10 epochs for NER.
After the first iteration, the model is trained for 3 epochs in each later iteration.
For LEU, we set the Monte Carlo sampling $T=20$.
For EVI, we set $\alpha = 1$ for NLI and $\alpha = 10^{-2}$ for NER based on the empirical scale of \textit{vac} and \textit{diss}, keeping both on the same scale.

To avoid the training set growing too huge as the SL process iterates, we apply a sampling strategy upon new selection: each training epoch samples from the existing training set with equal amount of newly selected data, so that each training epoch consists of at least 50\% latest selection.
We adopt early stop on English dev set if the evaluation does not improve for over two iterations.

Our experiments uses NVIDIA Titan RTX GPUs. The training takes 10 hours for both NER and NLI.

% Our code and models will be publicly available; code and running instructions are provided in the supplemental materials.

\subsection{Other Uncertainties}
\label{apdx:common}

\textbf{MPR} is the max probability of label classes, denoted by $\gamma = \max_c \mathbf{p}_c$. It is equivalent to the probability of the predicted label, and is commonly used as the selecting criterion for classification tasks \citep{sl1,sl2}.

\noindent \textbf{ENT} is the entropy of the class probability distribution, denoted by $\gamma = -\sum_c \mathbf{p}_c \cdot \log \mathbf{p}_c$, which is\LN another common uncertainty metric for classification \citep{pmlr-v80-depeweg18a,xiao_quantifying_2019}.